\documentclass[letterpaper]{article}

\usepackage{natbib,alifeconf}  
\usepackage{float}
\usepackage{graphics}
\usepackage{subcaption}

\title{Modeling Wildfire Perimeter Evolution using Deep Neural Networks}
\author{Maxfield E. Green$^{1}$, Karl Kaiser$^{1}$ \and Nat Shenton$^1$ \\
\mbox{}\\
$^1$University of Vermont, Burlington,VT 05401 \\
maxfield.green@uvm.edu} 

\begin{document}
\maketitle

\begin{abstract}

With the increased size and frequency of wildfire events worldwide, accurate real-time prediction of evolving wildfire fronts is a crucial component of firefighting efforts and forest management practices. We propose a wildfire spreading model that predicts the evolution of the wildfire perimeter in 24 hour periods. The fire spreading simulation is based on a deep convolutional neural network (CNN) that is trained on remotely sensed atmospheric and environmental time series data. We show that the model is able to learn wildfire spreading dynamics from real historic data sets from a series of wildfires in the Western Sierra Nevada Mountains in California. We validate the model on a previously unseen wildfire and produce realistic results that significantly outperform historic alternatives with validation accuracies ranging from 78\% - 98\%.

\end{abstract}

\section{Introduction}

In the past decade, over 68,000,000 acres have been burned by wildfires in the United States alone. One of the many consequences of this damage is economic, resulting in over \$5.1 billion costs in 
infrastructural damage repair \cite{verisk}. Additionally, there are enormous environmental, physical and public safety risks associated with wildfires in wilderness-urban interfaces. 

Building predictive models to aid in wildfire preparation and containment efforts
is increasingly important. With advances in computational resources, wildfire modeling has become a key component to successful forest management. Accurate simulations of wildfires
can inform best practices for forest management, as well as real time response to wildfire events. In the past ten years, wildfire modeling has grown from 
fully physical models to data driven models that leverage artificial intelligence and increased coverage of fire events.

Traditional physical models are
derived from the fundamental laws of physics and chemistry. They model coupled dynamics of the physical systems like diffusion, advection, radiation, etc. Often, these are in the form of sets of coupled partial differential equations and can be subject to issues of numerical instability, computational complexity and mechanistic results. Further, these types of models must be  recallibrated to extend to different areas. 

Current models used by the U.S. Forest Service such as FARSITE require rich data input that may depend on field agents physical documentation of ecology, this is demanding of time and human resources \cite{finney1998farsite}. Further, physical models such as \cite{ferragut2007numerical} have shown issues in their ability to be implemented numerically. Further, all physical models are based on explicit features based on data but not revised by data. Thus, there is a growing data set of recorded fires that capture spatial spreading that is not currently being used in practice. 

\section{Deep Learning and Fire Perimeter Prediction}

To reduce computation time, increase accuracy and leverage the advances in satellite imagery, recent work has modeled wildfire dynamics with machine learning or evolutionary strategies. This area has seen great success with increased accuracy of perimeter prediction from historic fires \cite{ZhengZhong2017Ffss}, \cite{radke2019firecast}. Crowley et al \cite{crowley} applied a set of reinforcement learning algorithms to learn spreading policies from satellite images within an agent based model. 

Radke et al proposed a deep neural network algorithm titled FireCast that predicts 24 hour wildfire perimeter evolution based on Satellite images and local historic weather \cite{radke2019firecast}. FireCast achieves a $20\%$ higher average accuracy compared to the Farsite model \cite{finney1998farsite} used in current practice. 

Deep learning models that leverage the abundance of remote sensing data available have made a recent impression on wildfire modeling \cite{radke2019firecast}, \cite{crowley}. Modern models are becoming capable of  maintaining the precision and accuracy of traditional physical methods while offering more flexibility towards learning different environmental regions and timescales, solving two of the problems that historic fire models have faced. Models discussed in the literature review largely suffer from inflexibility and require a large overhead to be tuned to multiple climates.

Further, applications of deep learning in atmospheric weather events such as precipitation \cite{rodrigues2018deepdownscale,booz2019deep,lin2018dynamic} have seen recent success. The recent work of Google AI Group \cite{kaae2020metnet} is particularly relevant to modeling wildfire spread over time. The MetNet algorithm  outperformed current state of the art fully physical models using only historic local atmospheric conditions and topography. Employing recursive layers to capture the temporal features of precipitation events and axial attention to encourage the model to focus on pertinent bounding boxes within the image based feature set, the model was able to learn high dimensional and robust features. MetNet serves as an example of the power of deep neural models to make generalized long term associations between remote sensing data and natural physical processes.

\subsection{Dataset}

\subsubsection{Geo data processing and data pipeline:}

The collection, processing, and use of geo data is necessarily complicated due to the many systems and standards needed to describe the real world and the difficulty that comes with unifying data from them. To streamline this process, a data pipeline was constructed to conduct three primary tasks; raw data collection, data preprocessing, and data sampling. To do this, heavy use of several United State government agency and commercial APIs was used to collect data, followed by a multistep process of cropping, geo-synchronizing, and reprojecting data layers into a final tensor to sample from. This process is summarized in Figure \ref{fig:pipeline} below:

\begin{figure}[h]
    \centering
    \includegraphics[scale=0.35]{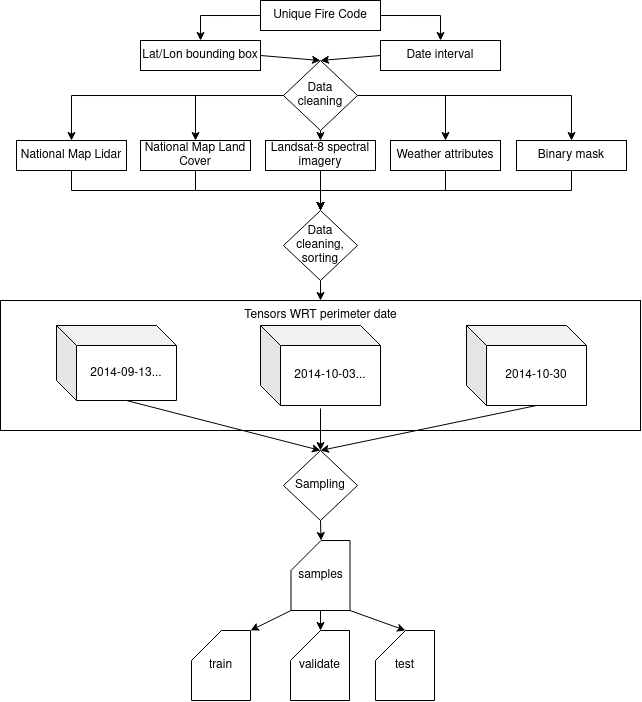}
    \caption{Summarized depiction of the data pipeline}
    \label{fig:pipeline}
\end{figure}

This process was conducted in a step-wise fashion. First, the Geomac historical wildfire perimeter archives \cite{geomac} were used to collect a series of spatially embedded and temporally distributed wildfire perimeter boundaries for a unique fire identifier. From these boundaries, a bounding box of maximum and minimum lat/lon coordinates and an overall date interval were extracted. Using this information, API calls were made to the API endpoints of the USGS Landsat 8 imagery archive \cite{earthexplorer}, the National Map viewer for both LiDAR and Land Cover \cite{usgs}, and to historical weather archives \cite{Meteostat}.

These APIs automatically collected the relevant data for whatever temporal resolution was possible within the time interval in which the wildfire occurred. The raw data fetching was followed by a significant amount of data cleaning, which involved geo-synchronizing the various data sources so that a pixel in one image was equivalent to a pixel in another. The primary type of data worked with were GeoTiffs and GeoJSONs of various coordinate systems, all of which were reprojected to depict a 30m $\times$ 30m pixel resolution under UTC (Universal Transa Mercator) coordinates. If wildfire perimeters spanned over multiple satellite snapshots, all necessary images required to represent the area upon which the fire spread were concatenated and cropped as needed, all with respect to the date time of the wildfire perimeter report. Finally, the images were stacked into tensors with respect to each wildfire perimeter report datetime, with each tensor using the most recent Landsat 8 spectral, LiDAR, and land cover imagery collected prior to the wildfire report. Finally, each tensor was capped with a binary mask of the wildfire's perimeter at that time. This method meant that any given tensor could contain satellite imagery captured up to one Landsat 8 rotational period (16 days) in advance of the relevant perimeter being collected, but never any satellite imagery after. If the USGS API reported no Landsat 8 imagery between 16 days prior to the perimeter and the perimeter itself, the perimeter was excluded.

Once tensors were constructed with respect to each perimeter date-time, they were filtered to include only one stack per 24 hour window (as perimeters could often times be reported within hours of each other). This was to ensure significant enough change between the binary masks of each tensor but also had the added advantage of reducing the size and complexity of the sampled data. Sampling was conducted under two schemes: binary classification / pixel prediction and mask generation. For both schemes, the sampled X data was an $n \times n$ slice of each layer. The corresponding Y data would either be the binary outcome of the pixel around which that sample was centered, or an $n \times n$ binary mask describing the area around the pixel around which the original sample was centered.

Due to the number of available samples and depth of the input tensors, the final data samples were both numerous and large in terms of their disk space requirements. To handle this, all samples were stored in a dynamic, non-compressed zipfile over which a custom generator could iterated to collect samples. The representation of outcomes were different for the two different sampling schemes; binary outcomes were embedded in each sample's filepath and split off from the generator during runtime. Mask outcomes were described in their own file with a filename corresponding to their relevant sample. A description of sampling within the context the experiment is available in the methodology sampling and Figure \ref{fig:sample_poi}.

\subsubsection{Produced Dataset:}
We have curated a dataset that covers a large span of wildfires in the Eastern Sierra of California, an area that is heavily effected every year by disastrous wildfires. This data was constructed entirely through use of the data pipeline described above and is well suited to be used as a training and testing set for learning models or other analyses of correlation between a given fire event's character and environmental variables present during the duration of that event. Perhaps one of the strongest attributes of this dataset is there is no unexpected stochastic in the samples collected from each wildfire's date-time relative data tensors, as both their random selection from the tensor and their subsequent shuffling into training / validation / testing sets is randomly seedable. 

This final dataset contained five fires: King (2014), Rocky (2015), Tubbs (2017), Cascade (2017), and County (2018). For each of these fires, we collected red, green, blue, and infared imagery from Landsat 8 archives \cite{earthexplorer}. Furthermore, we collected land cover (integer terrain classifications, used to distinguish rocky surfaces, vegetative surfaces, bodies of water, etc) and LiDAR data from the USGS National map viewer \cite{usgs}. Weather attributes such as wind speed and wind direction were described as single-valued matrices with values from the closest weather station. Finally, all tensors for each fire included a binary mask (fire, not on fire) of the wildfire perimeter at the relevant date-time. Specifics on how this data was sampled is available in the experimental methodology section.

\subsection{Algorithm Definition}
The model described in Table \ref{table:model1} is simple and relatively shallow compared to many popular computer vision learning models. The model was originally chosen to be simplistic to reduce training and testing times, as we were most interested in evaluating the applicability of the model to multiple, diverse fires rather than a single monolithic prediction task. However, this simple architecture ultimately yielded the best results.

\begin{table}[H]
\centering
\begin{tabular}{llll}
Layer  & Operation   & Kernel/Pool Size & Feature Maps \\
\hline
1      & Convolution & 7 x 7            & 128       \\
       & Max Pooling & 2 x 2            & -            \\
2      & Convolution & 6 x 6            & 64           \\
3      & Convolution & 3 x 3            & 128       \\

4      & Convolution & 3 x 3            & 256          \\
       & Max Pooling & 2x2              & -            \\
       & Flatten     & -                & -         \\

5      & Dense       & -                & 1024          \\
6      & Dense       & -                & 1024        \\
7-out & Dense       &                  & 1           
\end{tabular}
\caption{DNN model architecture}
\label{table:model1}
\end{table} 

The model takes in a sample with $31\times31\times8$ input dimensions, using the Adam optimizer with a learning rate of $0.0001$, and binary cross entropy as the loss function. The 8 input dimensions are then sampled from the stacked tensors described in the data pipeline. The model then predicts if the middle pixel will be on fire in $24$ hours. Thus, for each pixel position, the model is taking in the states (on fire or not on fire) and attributes (layers) of the surrounding areas, then computing the ultimate state of the center pixel at the next time step. If each pixel in an image represents a $30 \times 30$ meter square, then this model takes a (roughly) $961 \times 961$ meter square area and predicts if the the middle $30\times30$ meter square (pixel) will be on fire in 24 hours.  This model was implemented in Python 3, using the Keras and Tensorflow Deep Learning Frameworks. The model was tested against other more complex algorithms including known architectures such as ResNet50. This input/output scheme was chosen in part because the relevant subsampling methods allow use of relatively little data, and because this seemed like the ideal method to provide to the model as much local information as possible when considering the center pixel.

\section{Experimental Evaluation}

\subsection{Methodology}
\subsubsection{Stochastic Spatial Sampling:}
Sampling is performed without replacement, this way, it is impossible for the same location to be drawn twice from the same time step. Sampling without replacement is especially important to eliminate the chance of a data leak between testing and training sets. Samples are selected from a set of potential center pixels. Since this set contains all pixels within each date-time tensor, tensors are padded with 0 values on all layers. Thus, the model must employee a bit of guesswork for edge samples.  
 
\begin{figure}[H]
    \centering
    \includegraphics[scale=0.17]{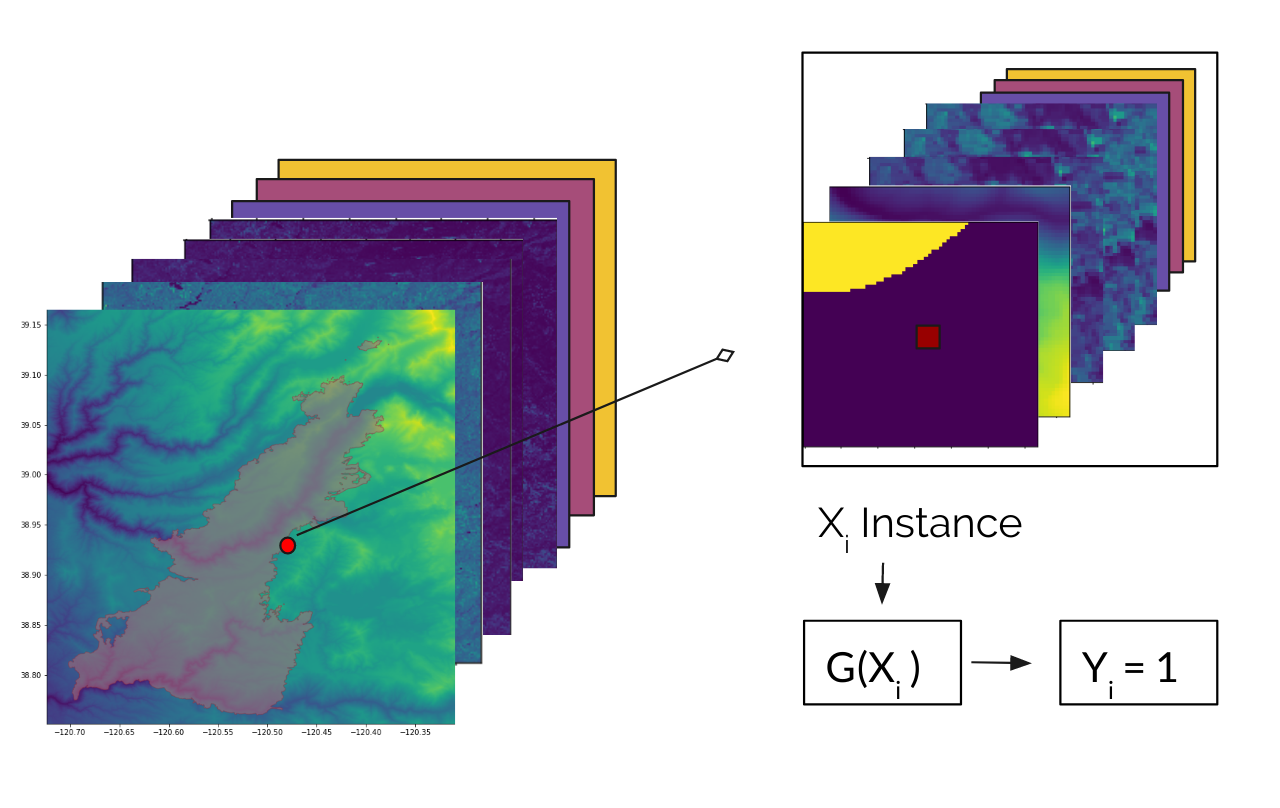}
    \caption{Example sample instance from the King Fire near Lake Tahoe CA. Here the $X_i$ instance shows state of the $31\times31$ grid with all 8 layers, and the historical weather data is placed at the input to the network $G(X_i)$ and in the $X_i$ instance in the figure the associated label is $Y_i=1$ as that instance will be with in the boundary in the next time step.}
    \label{fig:sample_poi}
\end{figure}

 We will refer to these sampled pixels as "points of interest" (POI). Once POI have been selected, their neighborhood is stored, as displayed visually in Figure \ref{fig:sample_poi}. For training and validation, a set of neighborhoods and their corresponding POI label in the next time step is sampled from every fire in the data set and from every available 24 hour period.  From each 24 hour tensor snapshot from a fire we sample 20,000 unique sub-tensors and corresponding labels.  As different fires burn for different durations, each fire has a different number of days the fires burned.
 
These neighborhoods are the input to a deep convolution neural network that is learning to predict the associated label. 

\subsection{Results}
For each of the fire simulations, we provide the accuracy and loss over 100 epochs in Figure \ref{fig:accuracy}. See Figure \ref{fig:loss_five} and Figure \ref{fig:acc_five} in Appendix for full result set.  Overall, we see positive performance. We notice that across the board, validation loss and accuracy is unstable, there are dips and rises in each of the learning curves. We see that this is dramatized by smaller data sets, as exemplified with the performance of the Cascade and Tubbs Fires. 


\begin{figure*}[!htb]
\centering

\minipage{0.48\textwidth}  
\includegraphics[width = \linewidth]{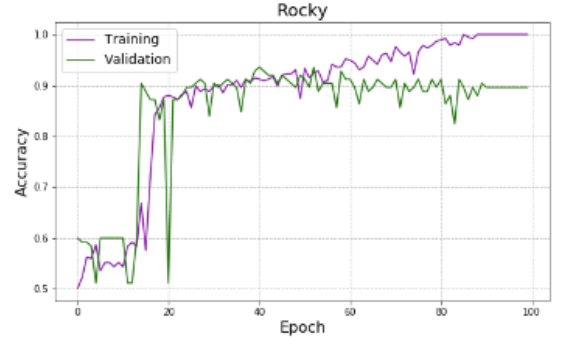}
 \endminipage\hfill
\minipage{0.48\textwidth}
\includegraphics[width = \linewidth]{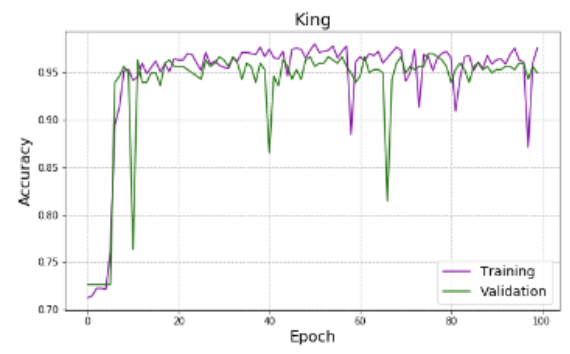}
  \endminipage

  \caption{The binary classification task accuracy for two sample fires.  Each fire was run for 100 epochs. The training accuracy does converge, and the model seems to only overfit for the Rocky and Cascade fires. See Fig \ref{fig:acc_five} in the appendix for total regional performance.}
  \label{fig:accuracy}
\end{figure*}
These results are consistent with the King, County, Tubbs, and Rocky fires as they all reach a validation accuracy over $90\%$ but the Cascade fire does not reach a validation accuracy over $80\%$.  The Cascade dataset is much small with only $\approx 40,000$ images after augmentation.  This is compared to the King fire with $\approx 240,000$ images after augmentation.  The Tubbs and Rocky fires have $\approx 100,000$ images after augmentation, and the County fire has $\approx 80,000$ images.  The algorithm is able to predict whether a pixel is on fire in the next time step depending on the $31\times31\times8$ picture. To understand the predicted class balances of each fire model, the precision, recall and accuracy from a validation sets are aggregated in Table  \ref{tab:model_results}. 
\begin{table}[]
\begin{tabular}{llll}
\centering
Fire    & Validation Accuracy & Recall & Precision \\
\hline
County  & 0.94                & 0.91   & 0.98      \\
King    & 0.97                & 0.97   & 0.98      \\
Rocky   & 0.91                & 0.88   & 0.96      \\
Tubbs   & 0.96                & 0.96   & 0.97     \\
Cascade & 0.77                & 0.75   & 0.82      
\end{tabular}
\caption{Descriptive results of model performance on validation set indicate that each model was able to predict the states of validation pixel neighborhoods with a high accuracy balanced across classes. }
\label{tab:model_results}
\end{table}

From Table \ref{tab:model_results}, we see a slight preference towards false negatives, with a lower recall than precision.  Meaning that the model is predicting that a pixel is not on fire when it actually is. Looking at the recall of this model, the Tubbs fire has the highest with $93.3\%$ while the Cascade has the lowest with $69.8\%$. The King fire has a recall of $90.9\%$, the County fire $87.3\%$ and the Rocky fire $85.1\%$. The results do vary with each fire. 

\subsubsection{Regional Model Training}
After pooling the data with all 5 fires, we train the binary architecture over $\approx 540,000$ images.  The model with the pooled data will give the sense of the predictions over the full region of fires rather than just a single one.  

\begin{figure}[H]
\centering
  \begin{subfigure}[t]{.48\textwidth}
  \includegraphics[width=\linewidth]{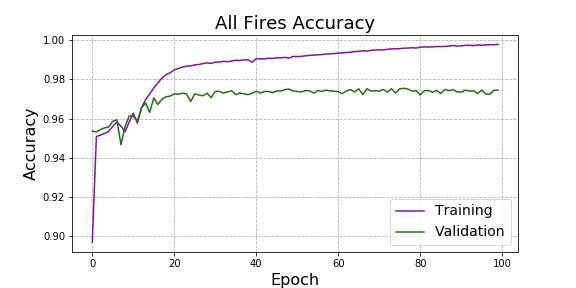}
  \caption{The accuracy per epoch of the train and validation sets for pooled fires.}
  \end{subfigure}
   \begin{subfigure}[t]{.48\textwidth}
  \includegraphics[width=\linewidth]{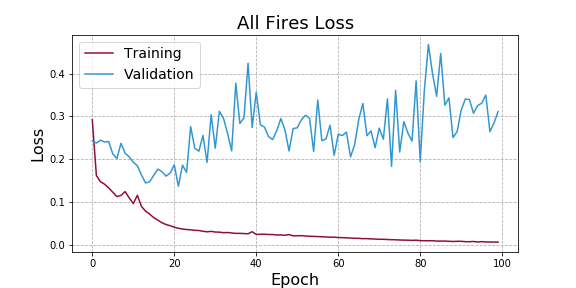}
  
  \caption{The binary loss per epoch of the train and validation sets for pooled fires.}
  \end{subfigure}
  \caption{}
  \label{fig:all_fires}
  
\end{figure}

\begin{figure}[H]
    \centering
    \includegraphics[width = 0.5\textwidth]{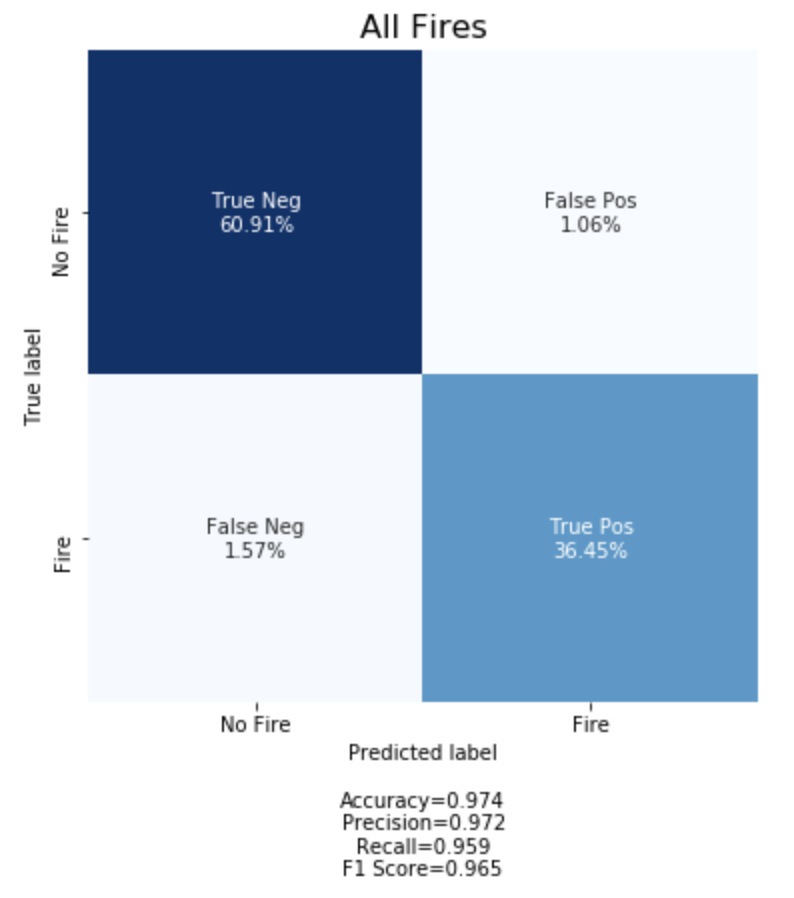}
    \caption{The confusion matrix of the pooled fires.  Each label is normalized based on the total number of samples in the test set. The final accuracy, precision. recall and F1 score is located below.}
    \label{fig:all_fire_con_matrix}
\end{figure}

Figure \ref{fig:all_fires} shows the loss and accuracy for training over 100 epochs.  The training does start to over fit after the first 15 epochs, and the validation accuracy does seem to plateau at about $97.5\%$ accuracy.  The loss follows the same trend at the accuracy as the model seems to start overfitting around epoch 15.  To understand the actually predictions of the model.  The confusion matrix with the testing set of $180,000$ images is shown in Figure \ref{fig:all_fire_con_matrix}.

Figure \ref{fig:all_fire_con_matrix} shows the model predicts false negative and false positive at a fairly event rate.  This is shown with the high precision, recall, and F1 scores.  The model is able to take a random $31\times31\times8$ tensor from the stack and predict if the middle pixel will be on fire with $97.4\%$ accuracy.  This regional model gives the a sense of possible predictions given the historical data of these fires.

\subsubsection{Regional Model Validation}
To validate the Regional Sierra Nevada fire, model we present an additional data set documenting the progression of a held out wildfire, specifically, the Rim Fire, which ran from August 17th, 2013 to November 4th of 2014, burning 257,6314 acres of land in the Sierra 
Nevada mountain range. To validate the model, we will sample the spatial data tensors as described in the training data generation and make 24 hour predictions of selected points across a subset of time intervals representing the initial, halfway point and final stages of the fire. 

We show that from the inputs of the previous days environmental conditions and fire perimeter, the model can accurately predict the following days across all sampled days of the wildfire.

\section{Conclusion}
We propose a CNN based regional wildfire forecasting model. By training the model to associate changes in the spatial atmospheric and environmental conditions, with changes the perimeter of the wildfire we accurately predict the evolution of several wildfires in the Sierra Nevada region of California. 
The spatial inputs include temporal red, blue, green and ultraviolet spectral images, land-cover and pixel state masks, temperature, wind speed, wind direction and precipitation. All data instances are temporally synchronized to the highest frequency possible given the availability of the raw data. The model takes in a 31x31x8 tensor and outputs a binary value indicating the future state of the tensors geo-located center. We show that the model is able to make accurate predictions on held out samples from training fires and completely unseen wildfires in similar regions within the Sierra Nevada's. The predictions hold up for different stages of the fire, showing both generality over different types of landscapes and temporal phases of the fire. 

\subsection{Next Steps}
The next steps of this work include building a general model using world wide recorded fires and applying transfer learning to fine tune the model to particular regions. Additionally, by performing classification on every single possible point within a region, the model will generate a continuous dense prediction mask allowing the predictions to be passed back into the model as a new set of initial conditions. 
\section{Acknowledgements}

This work was supported by Vermont Advanced Computing Core funded in part by NSF award No. OAC-1827314.

\bibliographystyle{apalike}
\bibliography{references} 

\begin{thebibliography}{}

\bibitem[Booz et~al., 2019]{booz2019deep}
Booz, J., Yu, W., Xu, G., Griffith, D., and Golmie, N. (2019).
\newblock A deep learning-based weather forecast system for data volume and
  recency analysis.
\newblock In {\em 2019 International Conference on Computing, Networking and
  Communications (ICNC)}, pages 697--701. IEEE.

\bibitem[Ferragut et~al., 2007]{ferragut2007numerical}
Ferragut, L., Asensio, M., and Monedero, S. (2007).
\newblock A numerical method for solving convection-reaction-diffusion
  multivalued equations in fire spread modelling.
\newblock {\em Advances in Engineering Software}, 38(6):366--371.

\bibitem[Finney, 1998]{finney1998farsite}
Finney, M.~A. (1998).
\newblock Farsite: Fire area simulator-model development and evaluation.
\newblock {\em Res. Pap. RMRS-RP-4, Revised 2004. Ogden, UT: US Department of
  Agriculture, Forest Service, Rocky Mountain Research Station. 47 p.}, 4.

\bibitem[Ganapathi~Subramanian and Crowley, 2018]{crowley}
Ganapathi~Subramanian, S. and Crowley, M. (2018).
\newblock Using spatial reinforcement learning to build forest wildfire
  dynamics models from satellite images.
\newblock {\em Frontiers in ICT}, 5:6.

\bibitem[Kaae~S{\o}nderby et~al., 2020]{kaae2020metnet}
Kaae~S{\o}nderby, C., Espeholt, L., Heek, J., Dehghani, M., Oliver, A.,
  Salimans, T., Agrawal, S., Hickey, J., and Kalchbrenner, N. (2020).
\newblock Metnet: A neural weather model for precipitation forecasting.
\newblock {\em arXiv}, pages arXiv--2003.

\bibitem[Lin et~al., 2018]{lin2018dynamic}
Lin, S.-Y., Chiang, C.-C., Li, J.-B., Hung, Z.-S., and Chao, K.-M. (2018).
\newblock Dynamic fine-tuning stacked auto-encoder neural network for weather
  forecast.
\newblock {\em Future Generation Computer Systems}, 89:446--454.

\bibitem[Meteostat, 2020]{Meteostat}
Meteostat (2020).
\newblock Meteostat statistical weather and climate data.
\newblock https://api.meteostat.net.
\newblock Accessed: 2020-05-01.

\bibitem[Radke et~al., 2019]{radke2019firecast}
Radke, D., Hessler, A., and Ellsworth, D. (2019).
\newblock Firecast: leveraging deep learning to predict wildfire spread.
\newblock In {\em Proceedings of the 28th International Joint Conference on
  Artificial Intelligence}, pages 4575--4581. AAAI Press.

\bibitem[Rodrigues et~al., 2018]{rodrigues2018deepdownscale}
Rodrigues, E.~R., Oliveira, I., Cunha, R., and Netto, M. (2018).
\newblock Deepdownscale: a deep learning strategy for high-resolution weather
  forecast.
\newblock In {\em 2018 IEEE 14th International Conference on e-Science
  (e-Science)}, pages 415--422. IEEE.

\bibitem[Sacadura, 2007]{verisk}
Sacadura, J.-F. (2007).
\newblock 2019 verisk wildfire risk analysis: Property underwriting.

\bibitem[USGS, 2020a]{earthexplorer}
USGS (2020a).
\newblock Earthexplorer.

\bibitem[USGS, 2020b]{geomac}
USGS (2020b).
\newblock Geomac database.

\bibitem[USGS, 2020c]{usgs}
USGS (2020c).
\newblock The national map: https://viewer.nationalmap.gov/advanced-viewer/.

\bibitem[Zheng et~al., 2017]{ZhengZhong2017Ffss}
Zheng, Z., Huang, W., Li, S., and Zeng, Y. (2017).
\newblock Forest fire spread simulating model using cellular automaton with
  extreme learning machine.
\newblock {\em Ecological Modelling}, 348:33--43.

\end{thebibliography}
\newpage
\newpage
\section{Appendix}
\begin{figure*}
    \centering
    \includegraphics[width = \linewidth]{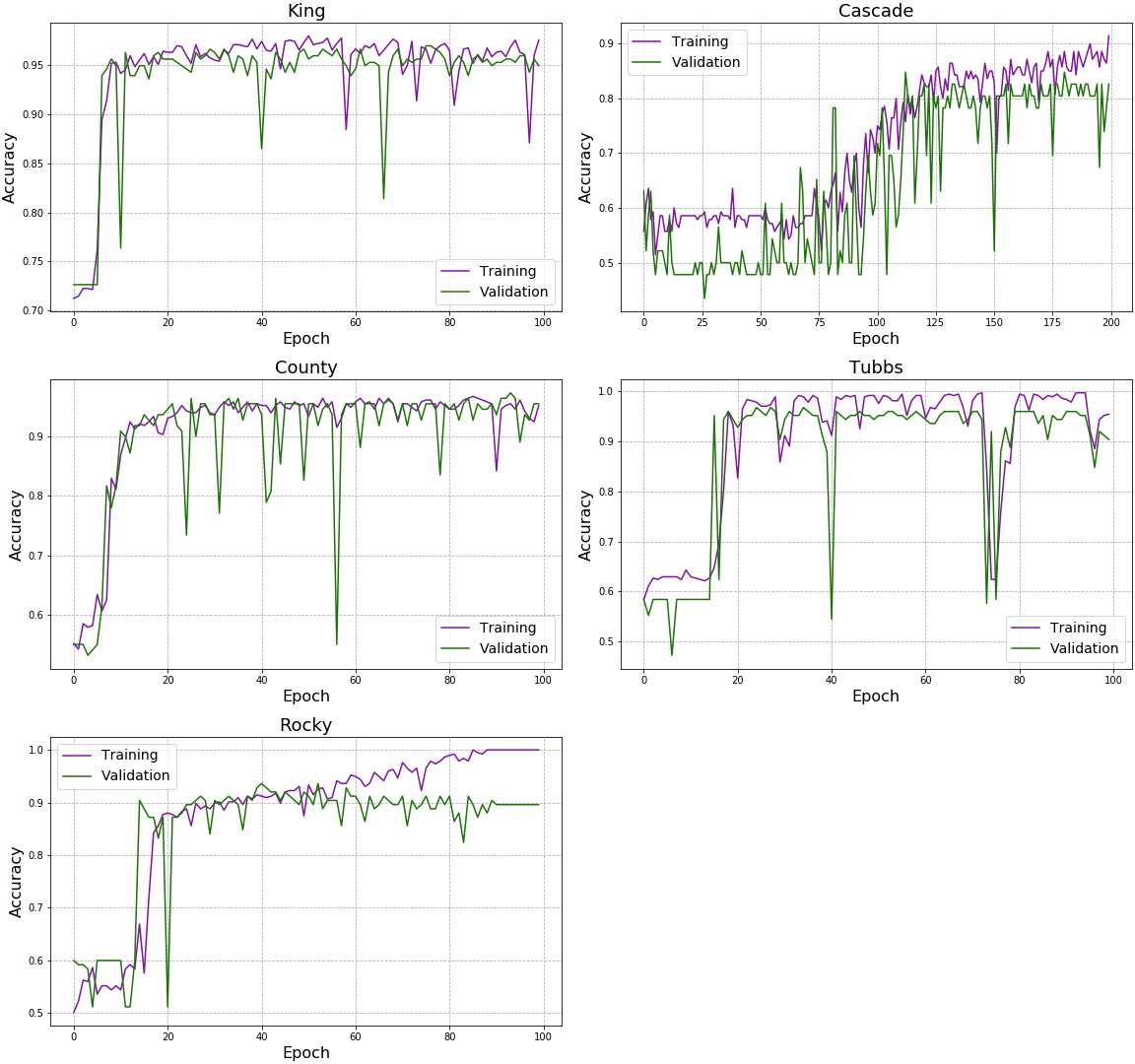}
    \caption{Individual fire model accuracies are steadily increasing with notable instability, potentially due to a relatively smaller data set. Notably the Tubbs and Cascade trained models show the largest variance and least convergence.}
    \label{fig:acc_five}
\end{figure*}

\begin{figure*}
    \centering
    \includegraphics[width = \linewidth]{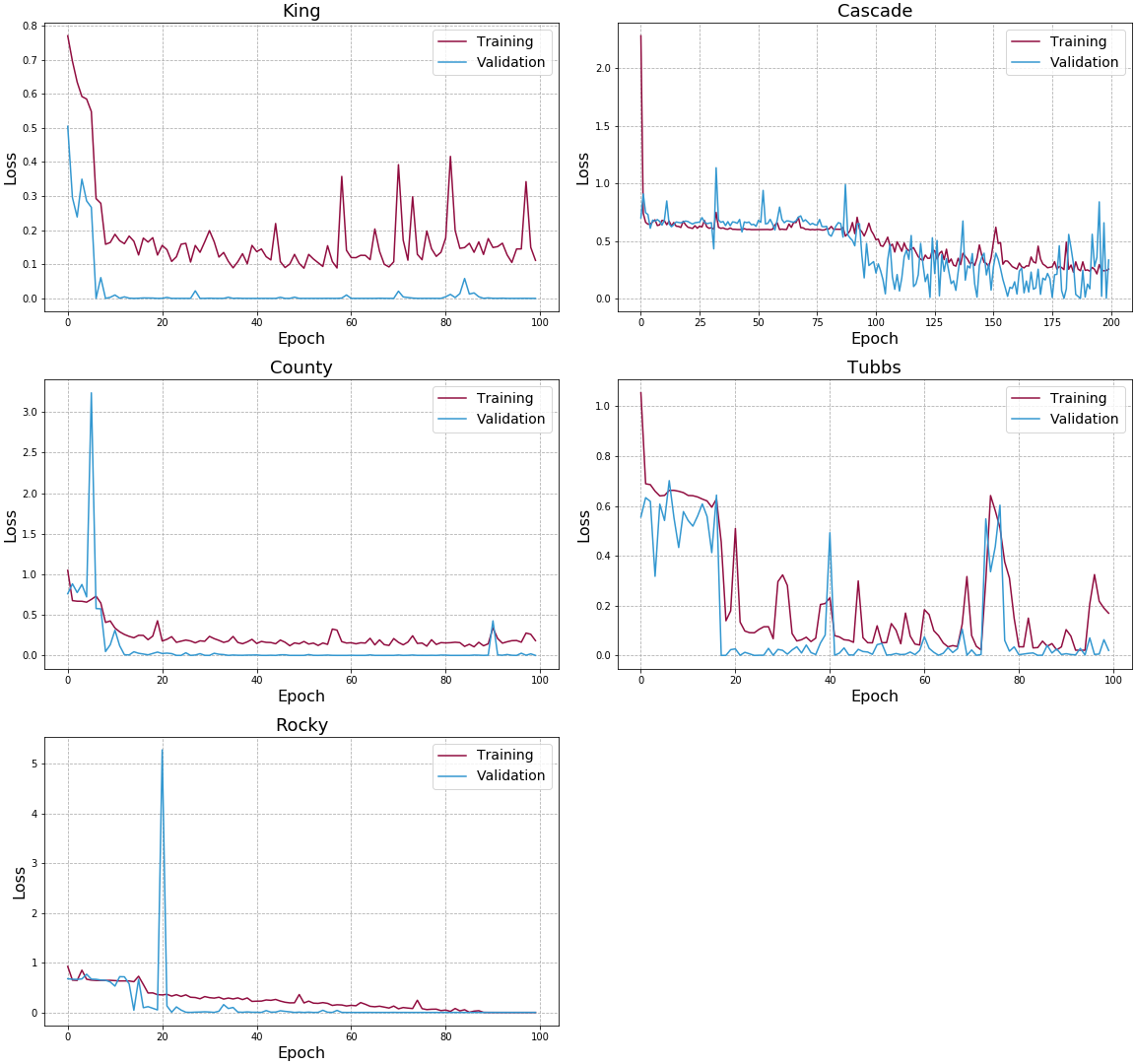}
    \caption{Caption}
    \label{fig:loss_five}
\end{figure*}

\end{document}